\documentclass[11pt,a4paper]{article}

\usepackage[utf8]{inputenc}
\usepackage[T2A,T1]{fontenc}
\usepackage[ukrainian,english]{babel}
\usepackage{amsmath,amssymb}
\usepackage{graphicx}
\usepackage{booktabs,multirow,array,tabularx}
\usepackage{hyperref}
\usepackage{natbib}
\usepackage[margin=1in]{geometry}
\usepackage{microtype}
\usepackage{enumitem}
\usepackage{caption}
\usepackage{pgfplots}
\pgfplotsset{compat=1.18}
\usepackage{float}
\usepackage{xcolor}

\newcolumntype{R}[1]{>{\raggedleft\arraybackslash}p{#1}}

\hypersetup{
    colorlinks=true,
    linkcolor=blue!70!black,
    citecolor=blue!70!black,
    urlcolor=blue!70!black,
    pdftitle={Gated Against One Model, Open to the Next: Option-Only Solvability in Legal Multiple-Choice Benchmarks},
    pdfauthor={Volodymyr Ovcharov},
    pdfsubject={benchmark validity, multiple-choice evaluation, legal NLP},
    pdfkeywords={benchmark validity, multiple-choice evaluation, shortcut learning, legal NLP, Ukrainian, contamination},
}

\title{Gated Against One Model, Open to the Next:\\
Option-Only Solvability in Legal Multiple-Choice Benchmarks}

\author{
  Volodymyr Ovcharov \\
  LEX AI \\
  \texttt{vladimir@legal.org.ua} \\
  \url{https://legal.org.ua}
}

\date{}

\begin{document}
\maketitle

\begin{abstract}
Multiple-choice benchmarks are graded on whether a model picks the right option, not on whether
it needed the question. Measuring that gap takes care: a model answering \texttt{A} to most
items scores above chance wherever the key sits at \texttt{A}, and reads as recognition when it
is not. We measure it on \textbf{UA-JudgeExam}: 11{,}990 four-option items
with official keys, published by Ukraine's Higher Qualification Commission of Judges.

Shown the options and \emph{no question}, Claude Haiku~4.5 scores $0.383$ against chance, and
the leak is concentrated: $11.8\%$ of items are answered blind on all eight option
orders, against $0.2$ items expected by chance. It is not quotation: search over $280{,}059$ editions
of Ukrainian legislation recovers $0.128$. Gating those out retains $8{,}128$ items, on
which the gating model itself now scores $0.204$ --- and GPT-5.6, which took no part in the
selection, still answers $0.515$ of them with the question hidden. Scoring eighteen held-out
models from eight vendors on the whole set and subtracting each one's position habit, four keep a
sizeable excess: GPT-5.6 ($+0.265$), Gemini~3.1~Pro ($+0.240$), GPT-5.4 ($+0.105$), Sonnet~4.6
($+0.081$); the rest stay within $0.031$, and probes find no recall of the public bank behind it.
Without the correction Llama~3.1~8B ($0.292$ blind, \texttt{A} on $92\%$ of items) would read as a leaker.

The gate does select something real: on the items it rejected, eleven of twelve models score
$0.518$--$0.789$, every interval clear of what the same model scores on the items it kept. But
that signal is one model's, and filtering on it does not transfer upward. Rewriting distractors instead overshoots
to $0.168$, below chance and as exploitable. The same probe on
LEXam returns chance: every option there points into the stem, none longer than 33 characters. Item format decides whether the problem can arise;
capability decides how much is extracted. We release the corpus, the predictions and the
harness.
\end{abstract}


\section{Introduction}

A multiple-choice benchmark reports one number: how often the model picks the key. That number
is treated as evidence about the model's competence in the domain. It is only such evidence to
the extent that the question is load-bearing --- that the model could not have found the key
from the options alone.

The obvious repair, when a bank fails that test, is to filter it: drop the items a model can
answer without the question and keep the rest. This paper's central finding is that the repair
does not hold. We filtered a state-published judicial examination bank until the filtering model
scored $0.204$ blind on what survived, below chance; GPT-5.6, which took no part in the
selection, still answers $0.515$ of those same items with the question hidden. A bank cleaned
against one model is not clean for a better one, and we found no version of the procedure that
made it so.

This is not a hypothetical concern wherever the options are self-contained propositions, as
they are in professional licensing material. Distractors are written by humans under time
pressure, and a wrong option is wrong for a reason: it names the wrong authority, states the
wrong deadline, or asserts something no statute says. A reader who knows the domain may then be
able to recognise the one option that reads like real law without ever seeing the question ---
and, as we show, so can a model, though we do not test human readers. To the extent that
happens, the benchmark measures recognition of well-formed legal propositions rather than the
reasoning it claims to measure. Where the options are instead pointers into
the stem --- ``i and iii'' --- the concern cannot arise, and we show that this distinction,
rather than subject matter, is what separates the benchmarks we test.

We make that quantity the object of study rather than a footnote. Our contributions:

\begin{enumerate}[leftmargin=*,itemsep=2pt]
\item \textbf{Filtering does not transfer.} Dropping the items one model can guess retains
$67.8\%$ of the bank and drives the gating model itself to $0.204$ on what survives --- yet
GPT-5.6 still answers $0.515$ of the cleaned set with the question hidden. Rewriting distractors
instead overshoots below chance. Neither repair produces a neutral set
(\S\ref{sec:vendors},~\S\ref{sec:failed}).
\item \textbf{A measurement protocol, and an estimator that separates habit from content.}
The blind condition is run over option-order permutations with the selecting model held separate
from the reporting one. Forcing the key into each slot in turn then makes the four accuracies
sum to one whenever choice is independent of content, so their mean is chance by construction
and any excess is content. Measured that way, ten of twelve held-out models extract at most $0.015$, five further models from four vendors repeat that split (two large excesses, three at most $0.031$), and neither prompt phrasing, option labelling nor inference-time reasoning accounts for
what the two leaking models of the first set extract (\S\ref{sec:gate},~\S\ref{sec:crossvendor},~\S\ref{sec:ablation}--\S\ref{sec:reasoning}).
\item \textbf{A benchmark that does not leak, and why.} The same probe on LEXam returns chance;
every one of its items uses reference-style options, none longer than 33 characters. Option
format, not subject matter, decides whether the problem can exist (\S\ref{sec:lexam}).
\item \textbf{A resource.} UA-JudgeExam: 11{,}990 items with official state-issued keys,
extraction verified by an independent second path, plus the gated $8{,}128$-item subset and
every prediction behind the tables (\S\ref{sec:corpus}).
\end{enumerate}

\section{Related Work}

\paragraph{Answering without the question.} The probe we use is not new. \citet{balepur2024artifacts}
prompt models with the choices alone on three MCQA datasets and beat a majority baseline in 11
of 12 cases, and find no evidence that this stems from memorisation, nor that priors over
individual choices fully explain it. \citet{cho2026choices} attack the same concern from the scoring side,
proposing a metric that isolates how much the question contributes to a model's decision. Both
establish that the phenomenon exists in general-domain multiple choice. We take it as given and ask
the question that follows for a practitioner holding a benchmark and reaching for the obvious
fix: how large is the effect on professional legal exam material, does filtering remove it, does
a filter built with one model protect against
another, and what property of an item set decides whether the problem arises at all.

\paragraph{Legal benchmarks.} Legal NLP benchmarks are predominantly English
\citep{guha2023legalbench,chalkidis2022lexglue,hendrycks2021cuad}, and LEXTREME
\citep{niklaus2023lextreme} broadens coverage to 24 languages, though its tasks are
classification and token-level extraction rather than multiple choice, so the blind condition does
not apply to it. The \citet{ostling2023cambridge} corpus is the closest analogue to our
resource on the legal and ethical side --- a national court corpus released under restrictions
worked out with institutional review, where we can release in full only because the bank is a
state document outside copyright. Exam-derived benchmarks inherit the item-writing conventions of
the professions that produced them, including the distractor-writing conventions this paper is
about: LEXam \citep{fan2026lexam}
builds 340 law exams into a reasoning benchmark, and \citet{katz2024gpt} evaluate against a
professional licensing examination directly. We measure the blind condition on LEXam's four-choice
split in \S\ref{sec:lexam} and find no leakage at all, for a reason that turns out to be about how
its options are written rather than about its subject matter.

\paragraph{Position and selection bias.} A separate line of work shows that models are sensitive
to \emph{where} an option sits. \citet{zheng2024robust} document a systematic preference for
particular option IDs across 20 models and trace it to token bias --- extra probability mass on
the tokens \texttt{A}/\texttt{B}/\texttt{C}/\texttt{D} themselves --- and debias by permuting
option contents; \citet{pezeshkpour2024sensitivity} show the same sensitivity to option order.
That literature and this paper measure different things which are easy to confuse in a
single-order experiment, since a model that merely favours one slot will also score above chance
whenever the key happens to sit there. \S\ref{sec:position} separates them by forcing the key
into each slot in turn, and finds the two orthogonal: the model with the strongest slot
preference in our set is not the one that leaks most. Our labelling sweep also bears on the
mechanism, since relabelling the options in Cyrillic or with digits leaves the aversion to the
first slot essentially unchanged.

\paragraph{Shortcuts and artifacts.} Work on annotation artifacts in natural language inference
established that models exploit hypothesis-only signal \citep{gururangan2018annotation,
poliak2018hypothesis}; the blind condition is the multiple-choice analogue. The closest study in
law is \citet{watson2026shortcut}, who ask whether legal judgment prediction forecasts anything
or merely reads outcome-revealing language left in the judgment, on 33{,}158 UK Employment
Tribunal claims. Their answer and ours diverge in a way worth stating. They remove the leaking
\emph{features} and the task survives: Macro-F1 falls negligibly, so a real signal was there
underneath. We remove the leaking \emph{items} and the benchmark does not survive in the same
sense --- the surviving set is clean for the model that selected it and not for a stronger one.
Deleting a cue from every instance and deleting the instances that carry it are not the same
operation, and only the first leaves the measurement intact.

\section{UA-JudgeExam}
\label{sec:corpus}

\subsection{Source}

Ukraine's Higher Qualification Commission of Judges (\foreignlanguage{ukrainian}{ВККС})
publishes the complete question bank used for the anonymous written testing of candidates for
appellate-court judgeships, together with the key for every item. We use the bank published by
Commission decision of 15~July~2024, No.~221/\foreignlanguage{ukrainian}{зп}-24: five documents
totalling 1{,}672 pages, covering general legal knowledge and the administrative, commercial,
criminal, and civil specialisations. Each item has exactly four options, of which the document
marks exactly one \foreignlanguage{ukrainian}{правильна} (correct) and three
\foreignlanguage{ukrainian}{неправильна} (incorrect).

The bank is published as an annex to a decision (\foreignlanguage{ukrainian}{рішення}) of the
Commission. Article~8(1)(3) of the Law of Ukraine \emph{On Copyright and Related Rights}
(No.~2811-IX) places outside copyright protection ``acts of state authorities \ldots\ official
documents of a political, legislative, administrative and judicial character (laws, decrees,
resolutions, \emph{decisions}, state standards and the like)''. On that basis we redistribute
the bank with attribution to the issuing decision. The items are questions on points of law and
contain no personal data.

\subsection{What an item looks like}

\begin{table}[H]
\centering\small
\caption{Item \texttt{vkks-2024-commercial-2237}, with an English gloss; the key, C, is in bold.
The options are self-contained legal propositions, which is the property that makes the blind
condition meaningful. Article~31 of the Civil Code grants a person under 14 the right
\foreignlanguage{ukrainian}{\emph{самостійно вчиняти дрібні побутові правочини}} --- so D
substitutes \foreignlanguage{ukrainian}{\emph{різні}} (``various'') for
\foreignlanguage{ukrainian}{\emph{дрібні}} (``minor''), which is not legal language, and A
substitutes \foreignlanguage{ukrainian}{\emph{майнові}} (``property'') for the
\foreignlanguage{ukrainian}{\emph{особисті немайнові}} (``personal non-property'') rights the
article actually grants. Telling these apart requires knowing the code, not reading the
question.}
\label{tab:example}
\begin{tabular}{@{}p{0.47\textwidth}p{0.47\textwidth}@{}}
\toprule
\multicolumn{2}{@{}l}{\textbf{Q.} \foreignlanguage{ukrainian}{Які дії має право самостійно
вчиняти фізична особа, яка НЕ досягла 14 років?}} \\
\multicolumn{2}{@{}l}{\emph{Which acts may a natural person under 14 perform independently?}} \\
\midrule
\foreignlanguage{ukrainian}{A) Здійснювати майнові права на результати інтелектуальної
діяльності, що охороняються законом} & \emph{Exercise property rights in protected
intellectual-property results} \\
\foreignlanguage{ukrainian}{B) Розпоряджатися банківським вкладом, унесеним нею на своє ім'я}
& \emph{Dispose of a bank deposit made in their own name} \\
\textbf{\foreignlanguage{ukrainian}{C) Вчиняти дрібні побутові правочини}}
& \textbf{\emph{Perform minor everyday transactions}} \\
\foreignlanguage{ukrainian}{D) Вчиняти різні побутові правочини} & \emph{Perform various
everyday transactions} \\
\bottomrule
\end{tabular}
\end{table}

This is the format that leaks. Contrast it with LEXam (\S\ref{sec:lexam}), whose four-choice
items offer options like ``i und iii'' --- pointers into a list of statements given in the stem,
which carry no legal content of their own.

\subsection{Extraction and its verification}

The documents are ruled tables in PDF. We extract them with a table parser and verify the
result with a \emph{second, independent} path that reconstructs rows from the table's
horizontal ruling lines, columns from its vertical ones, and cell contents from word
coordinates --- sharing no code with the first. Comparison is insensitive to whitespace and
line-break hyphenation, because the two paths join wrapped words differently. That insensitivity
hid one defect of the first path: at the table parser's default horizontal tolerance, word spaces
are lost in 489 items ($4.1\%$, 354 of them in the criminal set), producing tokens such as
\foreignlanguage{ukrainian}{Заякихобставинвиконаннявійськовогообов'язку}. A tolerance of $1.5$\,pt
restores them without changing a single non-whitespace character or key. The defect falls evenly
on keys and distractors ($0.91\%$ against $0.84\%$ of options), so it is not a cue, and every
number in this paper is computed on the text as first released; we release the re-extracted
text alongside it.

Extraction yields \textbf{11{,}990} well-formed items, plus 385 items the Commission itself
marks as withdrawn (\foreignlanguage{ukrainian}{Виключено рішенням}) and 40 that fail the
four-options-one-key invariant. Those 40 are extraction failures rather than defects of the
source: the cell giving an option's type is letter-spaced in the PDF
(\foreignlanguage{ukrainian}{н е п р а в и л ь н а}), the parser does not recognise it and drops the
option, and a re-extraction that normalises the cell recovers all 40 with their keys. On a 200-item stratified pilot the second path confirms the
question text, the four options in order, and the key in $0.995$ of items. Over the full bank
it confirms the question text in $0.9896$, option order in $0.9888$, and \textbf{the answer key
in $0.9841$}; the 192 items ($1.6\%$) that fail at least one check are listed by identifier in
the release.

Five items are malformed in the source itself and we ship them as published rather than
repairing them silently: in one the fourth option is blank in the PDF, and four carry two
identical options. None of them falls in the 400-item sample used for the probes below, which
contains no empty options; two of the four with identical options are among the $8{,}128$ accepted
items.

One extraction detail is worth reporting because it is a silent failure mode: in 942 items the
cell holding the item number does not extract, which naively merges two adjacent items into one
eight-option record. Detecting the merge by the appearance of a new question stem while options
are already accumulating recovers those items; the recovered numbering contains zero duplicates
within any of the five documents, which is the namespace the Commission numbers in and the
check that the recovery is sound.

\section{The Solvability Profile}
\label{sec:profile}

Before asking what models score, we establish what trivial strategies score. Unless noted, all
numbers are over the full bank of 11{,}990 items, on the options as published.

\begin{table}[H]
\centering
\caption{What can be recovered without reasoning over the question. Chance is $0.250$.}
\label{tab:profile}
\begin{tabular}{lr}
\toprule
\textbf{Strategy} & \textbf{Accuracy} \\
\midrule
Chance & 0.250 \\
Best constant position (always A) & 0.289 \\
Longest option & 0.278 \\
Most words & 0.301 \\
Shortest option & 0.253 \\
Lexical search over statutes, full bank & 0.128 \\
Lexical search over statutes, items with four substantive options ($n{=}8{,}024$) & 0.172 \\
\midrule
\emph{Blind LLM, options only} (Haiku 4.5, 8 permutations) & \textbf{0.383} \\
\bottomrule
\end{tabular}
\end{table}

Surface features carry almost nothing: the correct option is on average $2.6$ characters longer
than a distractor, and no length or position heuristic clears $0.301$. The key sits at A in
$28.9\%$ of items, at B and at C in $24.8\%$ each and at D in $21.5\%$, so a positional habit
buys a little; we control for it explicitly in \S\ref{sec:vendors}.

\subsection{Lexical search is not the mechanism}

A plausible story is that the correct option is copied from a statute while distractors are
invented, so a text search decides the item. We tested this against the full text of
\textbf{280{,}059 current editions} of Ukrainian legislation ($2.356$ billion characters). For
each option we ask whether its first twelve tokens occur verbatim anywhere in that corpus.

Over the full bank the correct option is found verbatim in $0.580$ of items and distractors in
$0.438$ --- a real but weak signal. The decisive statistic is what a searcher can do with it:
exactly one option is found in $19.6\%$ of items, and there the match is the key $65.1\%$ of the
time, for an overall accuracy of $0.128$. One- and two-token options --- bare figures such as
``1095'' --- occur verbatim $94.6\%$ of the time ($10{,}637$ of $47{,}960$ options). They
inflate the raw match rates and, because they match for every option at once, \emph{depress}
what a searcher can do with them, so we also restrict to the $8{,}024$ items whose four options
are all substantive (three tokens or more). There the
searcher commits on $25.9\%$ of items and is right $66.4\%$ of the time when it does, for
$0.172$ overall --- still below chance.

We ran this two ways: as phrase queries against a full-text index, and as a single pass over an
exported corpus with an inverted index of the option phrases. The two implementations, on
different machines with no shared code, agree on $97.11\%$ of individual option lookups over a
shared 1{,}000-item sample. They do not agree exactly on what a searcher gets: the Postgres
route is the more conservative, at $0.112$ against $0.136$ on those items. That gap is not the
$128$ timed-out Postgres queries: dropping every item they touch leaves it at $0.104$ against
$0.127$. It is a genuine recall difference, $109$ phrases found only by the offline index
against $3$ found only by Postgres. We report the higher of the two throughout, so the figure
above is the one least favourable to our argument.

Critically, blind accuracy does \emph{not} concentrate on items whose key is verbatim statutory
text. On the 199 items of the pilot for which both paths ran, Haiku scores $0.433$ on verbatim
keys against $0.347$ on the rest ($n=97$ and $75$; it returned no parseable letter on 27 of the
199) and Sonnet $0.366$ against $0.414$ ($n=112$ and $87$) --- the two models disagree even on
the direction, and every interval overlaps. The leak is not quotation. It is
plausibility.

\section{The Blind Gate}
\label{sec:gate}

\subsection{Protocol}

For each item we present the four options with the question withheld and ask for a single
letter, repeating over eight random permutations of option order. Under the null of no leakage
each trial is correct with probability $0.25$, so eight trials give a binomial reference:
$P(X \ge 5) = 0.027$. We reject an item at five or more hits, and additionally require at least
six of the eight trials to have produced a parseable letter, so that an item is never certified
clean on the strength of two or three responses. We do not reject on the lower tail --- zero
hits out of eight occurs for $10.0\%$ of clean items --- and treat below-chance behaviour as a
set-level diagnostic instead.

Two caveats on that reference. The eight orders are drawn independently rather than sampled
without replacement, so an item sees $6.9$ distinct orders on average and all eight only $27\%$
of the time; at temperature 0 a repeated order repeats its answer, which makes the trials
slightly less independent than the binomial assumes. Recomputing the tail against seven
effective trials moves the expected number of items answered blind eight times out of eight
from $0.2$ to $0.7$, against the $1{,}419$ observed, so the conclusion is unaffected and the
threshold is marginally conservative. Separately, the run did not persist per-item hit counts
and seeded from Python's salted string hash, so the histogram in Figure~\ref{fig:hits} is
recoverable from our logs but the selection is not bit-for-bit reproducible from the release;
the accepted set itself is published in full.

Two design points matter for the numbers to mean anything. First, the gate runs on the item as
published, not on a rewritten variant. Second, \textbf{the model that selects is not the model
that reports}: gating uses Haiku~4.5 and every reported blind figure comes from models that
took no part in selection.

\subsection{Results}

Over the full bank the pooled blind rate during gating is $0.383$. The distribution of hits per
item is strongly bimodal: 3{,}517 items ($29.3\%$) are never answered blind in eight attempts,
while 1{,}419 ($11.8\%$) are answered blind in all eight --- an event with probability
$1.5\times10^{-5}$ under chance. Leakage is concentrated in a minority of items, which is what
makes selection a plausible remedy at all.

\begin{figure}[H]
\centering
% Created by tikzDevice version 0.12.6 on 2026-08-15 14:44:42
% !TEX encoding = UTF-8 Unicode
\begin{tikzpicture}[x=1pt,y=1pt]
\definecolor{fillColor}{RGB}{255,255,255}
\path[use as bounding box,fill=fillColor,fill opacity=0.00] (0,0) rectangle (390.26,209.58);
\begin{scope}
\path[clip] (  0.00,  0.00) rectangle (390.26,209.58);
\definecolor{fillColor}{RGB}{255,255,255}

\path[fill=fillColor] (  0.00,  0.00) rectangle (390.26,209.58);
\end{scope}
\begin{scope}
\path[clip] ( 33.14, 25.11) rectangle (385.76,205.08);
\definecolor{drawColor}{gray}{0.92}

\path[draw=drawColor,line width= 0.5pt,line join=round] ( 33.14, 25.11) --
	(385.76, 25.11);

\path[draw=drawColor,line width= 0.5pt,line join=round] ( 33.14, 68.14) --
	(385.76, 68.14);

\path[draw=drawColor,line width= 0.5pt,line join=round] ( 33.14,111.17) --
	(385.76,111.17);

\path[draw=drawColor,line width= 0.5pt,line join=round] ( 33.14,154.20) --
	(385.76,154.20);

\path[draw=drawColor,line width= 0.5pt,line join=round] ( 33.14,197.23) --
	(385.76,197.23);
\definecolor{fillColor}{RGB}{44,93,143}

\path[fill=fillColor] ( 42.92, 25.11) rectangle ( 55.57,176.44);

\path[fill=fillColor] ( 81.25, 25.11) rectangle ( 93.89, 85.78);

\path[fill=fillColor] (119.57, 25.11) rectangle (132.22, 76.57);

\path[fill=fillColor] (157.90, 25.11) rectangle (170.55, 70.16);

\path[fill=fillColor] (196.23, 25.11) rectangle (208.88, 66.72);

\path[fill=fillColor] (234.56, 25.11) rectangle (247.20, 62.24);

\path[fill=fillColor] (272.88, 25.11) rectangle (285.53, 60.69);

\path[fill=fillColor] (311.21, 25.11) rectangle (323.86, 57.12);

\path[fill=fillColor] (349.54, 25.11) rectangle (362.19, 86.17);
\definecolor{fillColor}{gray}{0.79}

\path[fill=fillColor] ( 56.72, 25.11) rectangle ( 69.36, 76.76);

\path[fill=fillColor] ( 95.04, 25.11) rectangle (107.69,162.84);

\path[fill=fillColor] (133.37, 25.11) rectangle (146.02,185.80);

\path[fill=fillColor] (171.70, 25.11) rectangle (184.35,132.24);

\path[fill=fillColor] (210.03, 25.11) rectangle (222.67, 69.74);

\path[fill=fillColor] (248.35, 25.11) rectangle (261.00, 37.01);

\path[fill=fillColor] (286.68, 25.11) rectangle (299.33, 27.09);

\path[fill=fillColor] (325.01, 25.11) rectangle (337.66, 25.30);

\path[fill=fillColor] (363.34, 25.11) rectangle (375.98, 25.11);
\definecolor{drawColor}{RGB}{44,93,143}

\node[text=drawColor,anchor=base east,inner sep=0pt, outer sep=0pt, scale=  0.71] at (345.51, 98.39) {1,419 items answered blind every time;};

\node[text=drawColor,anchor=base east,inner sep=0pt, outer sep=0pt, scale=  0.71] at (345.51, 89.78) {chance predicts 0.2};
\end{scope}
\begin{scope}
\path[clip] (  0.00,  0.00) rectangle (390.26,209.58);
\definecolor{drawColor}{gray}{0.30}

\node[text=drawColor,anchor=base east,inner sep=0pt, outer sep=0pt, scale=  0.72] at ( 29.09, 22.63) {0};

\node[text=drawColor,anchor=base east,inner sep=0pt, outer sep=0pt, scale=  0.72] at ( 29.09, 65.66) {1000};

\node[text=drawColor,anchor=base east,inner sep=0pt, outer sep=0pt, scale=  0.72] at ( 29.09,108.69) {2000};

\node[text=drawColor,anchor=base east,inner sep=0pt, outer sep=0pt, scale=  0.72] at ( 29.09,151.72) {3000};

\node[text=drawColor,anchor=base east,inner sep=0pt, outer sep=0pt, scale=  0.72] at ( 29.09,194.75) {4000};
\end{scope}
\begin{scope}
\path[clip] (  0.00,  0.00) rectangle (390.26,209.58);
\definecolor{drawColor}{gray}{0.30}

\node[text=drawColor,anchor=base,inner sep=0pt, outer sep=0pt, scale=  0.72] at ( 56.14, 16.10) {0};

\node[text=drawColor,anchor=base,inner sep=0pt, outer sep=0pt, scale=  0.72] at ( 94.47, 16.10) {1};

\node[text=drawColor,anchor=base,inner sep=0pt, outer sep=0pt, scale=  0.72] at (132.80, 16.10) {2};

\node[text=drawColor,anchor=base,inner sep=0pt, outer sep=0pt, scale=  0.72] at (171.12, 16.10) {3};

\node[text=drawColor,anchor=base,inner sep=0pt, outer sep=0pt, scale=  0.72] at (209.45, 16.10) {4};

\node[text=drawColor,anchor=base,inner sep=0pt, outer sep=0pt, scale=  0.72] at (247.78, 16.10) {5};

\node[text=drawColor,anchor=base,inner sep=0pt, outer sep=0pt, scale=  0.72] at (286.11, 16.10) {6};

\node[text=drawColor,anchor=base,inner sep=0pt, outer sep=0pt, scale=  0.72] at (324.43, 16.10) {7};

\node[text=drawColor,anchor=base,inner sep=0pt, outer sep=0pt, scale=  0.72] at (362.76, 16.10) {8};
\end{scope}
\begin{scope}
\path[clip] (  0.00,  0.00) rectangle (390.26,209.58);
\definecolor{drawColor}{RGB}{0,0,0}

\node[text=drawColor,anchor=base,inner sep=0pt, outer sep=0pt, scale=  0.90] at (209.45,  6.25) {Blind hits out of 8 permutations};
\end{scope}
\begin{scope}
\path[clip] (  0.00,  0.00) rectangle (390.26,209.58);
\definecolor{drawColor}{RGB}{0,0,0}

\node[text=drawColor,rotate= 90.00,anchor=base,inner sep=0pt, outer sep=0pt, scale=  0.90] at ( 10.70,115.09) {Items};
\end{scope}
\begin{scope}
\path[clip] (  0.00,  0.00) rectangle (390.26,209.58);
\definecolor{fillColor}{RGB}{255,255,255}

\path[fill=fillColor] (246.53,166.28) rectangle (327.52,193.49);
\end{scope}
\begin{scope}
\path[clip] (  0.00,  0.00) rectangle (390.26,209.58);
\definecolor{fillColor}{RGB}{44,93,143}

\path[fill=fillColor] (251.61,180.47) rectangle (259.55,188.41);
\end{scope}
\begin{scope}
\path[clip] (  0.00,  0.00) rectangle (390.26,209.58);
\definecolor{fillColor}{gray}{0.79}

\path[fill=fillColor] (251.61,171.36) rectangle (259.55,179.30);
\end{scope}
\begin{scope}
\path[clip] (  0.00,  0.00) rectangle (390.26,209.58);
\definecolor{drawColor}{RGB}{0,0,0}

\node[text=drawColor,anchor=base west,inner sep=0pt, outer sep=0pt, scale=  0.72] at (264.64,181.96) {Observed};
\end{scope}
\begin{scope}
\path[clip] (  0.00,  0.00) rectangle (390.26,209.58);
\definecolor{drawColor}{RGB}{0,0,0}

\node[text=drawColor,anchor=base west,inner sep=0pt, outer sep=0pt, scale=  0.72] at (264.64,172.85) {Chance (binomial)};
\end{scope}
\end{tikzpicture}
\caption{How often each item is answered blind, over eight permutations of option order,
against what chance predicts. The bank does not behave like a set of items with a uniform
small leak: 3{,}517 items are never answered blind and 1{,}419 are answered blind every time,
where $\mathrm{Binomial}(8, 0.25)$ predicts $0.2$ items in the latter group. That concentration
is what makes selection worth attempting.}
\label{fig:hits}
\end{figure}
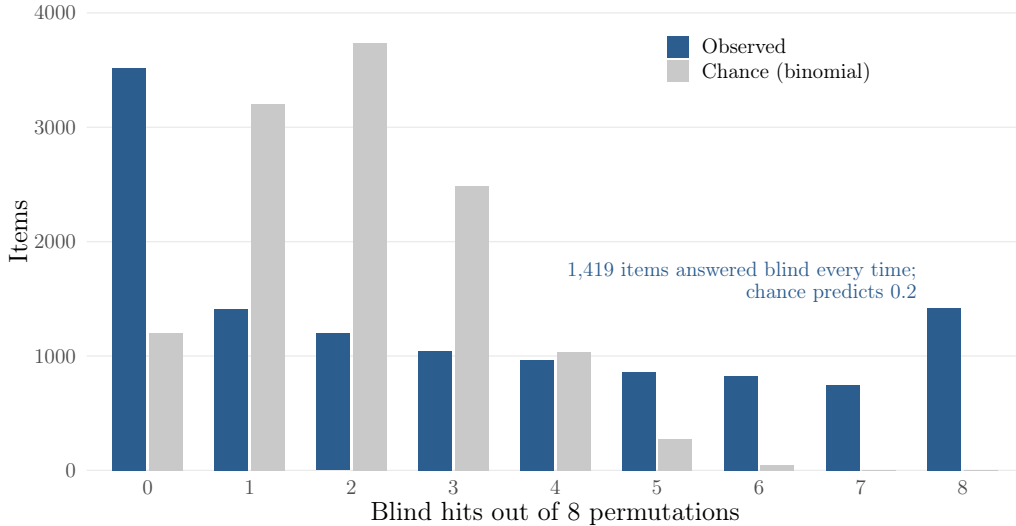

Figure~\ref{fig:hits} shows the shape. Of the $8{,}137$ items at four hits or fewer, nine fall
to the minimum-trials rule, so the gate retains \textbf{8{,}128 items ($67.8\%$)}. Retention
varies by specialisation:
administrative $74.3\%$, commercial $70.6\%$, civil $64.6\%$, criminal $63.2\%$, general
$61.4\%$.

On those $8{,}128$ items the gating model now scores $0.204$ blind. By its own measure the bank
is clean. Section~\ref{sec:vendors} asks what that measure is worth to anyone else, and the
answer is the reason this paper exists.

\section{Does the Repair Transfer?}
\label{sec:vendors}

The gate was built with a single model. We test whether the items it kept are clean for anyone
else, on a 600-item sample stratified by gate outcome --- 400 accepted and 200 rejected ---
scored by thirteen models from eight vendors under a uniform protocol. The sample is not a random
draw from the bank, so the figures below are conditional on gate outcome rather than bank-wide
averages. All models were reached through Amazon Bedrock in August 2026 with greedy decoding wherever the
provider accepts it. The blind condition on accepted items was subsequently re-run on the
\emph{entire} accepted set with a 4{,}096-token budget; the two runs agree closely where they
overlap. On the items both scored, ten of thirteen models differ by less than $0.013$ and the
largest difference is Pixtral's $0.017$ on the $233$ items it shares; Sonnet and Haiku return
identical predictions on all $400$. The differences between the columns below are therefore
sampling, not protocol. Pixtral Large is quota-bound on its inference profile and was
re-run on the whole accepted set. One warning for anyone subsampling the released file: the
bank is stored in the order of the five source documents, and the documents \emph{are} the
specialisations, so its first $2{,}000$ items are $74\%$ administrative. Stratify rather than
slice.

\begin{table}[H]
\centering
\small
\caption{Blind (options only) and full accuracy, ordered by full-condition accuracy. Chance
is $0.250$. \textbf{Blind figures on the accepted set are measured on all $8{,}128$ items};
the rejected-set and full-condition columns remain on the 600-item sample, so their intervals
are the wider ones. Two models are set apart because their numbers are not evidence
of recognition: Haiku is the gate's own model, and Llama~3.1~8B answers \texttt{A} to $92\%$ of
items, which on its own earns $0.287$ here (Table~\ref{tab:habit}). Blind $n$ is $8{,}128$ except DeepSeek~R1 $7{,}976$,
Llama~3.1~8B $7{,}886$, Nova~Micro $8{,}039$, Ministral $8{,}126$ and Pixtral $8{,}125$; full
$n$ is $400$ except DeepSeek~R1 $352$, Pixtral $365$, Nova~Micro $388$ and Llama~3.1~8B $202$.}
\label{tab:vendors}
\begin{tabular}{llR{2.9cm}R{2.9cm}rr}
\toprule
\textbf{Model} & \textbf{Vendor} & \textbf{Blind, accepted} & \textbf{Blind, rejected}
& \textbf{Full acc.} & \textbf{Full rej.} \\
\midrule
GPT-5.6        & OpenAI    & \textbf{0.515} [.504,.526] & 0.789 [.727,.840] & 0.958 & 0.960 \\
Sonnet 4.6     & Anthropic & \textbf{0.320} [.310,.330] & 0.730 [.665,.787] & 0.740 & 0.905 \\
DeepSeek R1    & DeepSeek  & 0.257 [.247,.266]          & 0.665 [.593,.729] & 0.668 & 0.845 \\
Nova Pro       & Amazon    & 0.232 [.223,.241]          & 0.645 [.577,.708] & 0.580 & 0.739 \\
Qwen3 32B      & Alibaba   & 0.244 [.234,.253]          & 0.665 [.597,.727] & 0.552 & 0.720 \\
Pixtral Large  & Mistral   & 0.239 [.230,.248]          & 0.602 [.531,.669] & 0.548 & 0.718 \\
Llama 3.3 70B  & Meta      & 0.237 [.228,.247]          & 0.610 [.541,.675] & 0.540 & 0.790 \\
Gemma 3 12B    & Google    & 0.230 [.221,.239]          & 0.595 [.526,.661] & 0.472 & 0.745 \\
Ministral 8B   & Mistral   & 0.239 [.229,.248]          & 0.540 [.471,.608] & 0.463 & 0.740 \\
Nova 2 Lite    & Amazon    & 0.225 [.216,.234]          & 0.545 [.476,.613] & 0.460 & 0.660 \\
Nova Micro     & Amazon    & 0.233 [.224,.242]          & 0.518 [.448,.586] & 0.436 & 0.658 \\
\midrule
Llama 3.1 8B   & Meta      & \emph{0.292} [.282,.302]   & 0.337 [.274,.406] & 0.426 & 0.531 \\
Haiku 4.5      & Anthropic & \emph{0.204} [.195,.213]   & 0.800 [.739,.850] & 0.575 & 0.805 \\
\bottomrule
\end{tabular}
\end{table}

Four readings follow, and the first is the answer to the question in the heading.

\paragraph{A gate built with one model does not protect against a better one.} The gate ran
until Haiku~4.5 scored $0.204$ blind on the items it kept --- below chance, its own leakage
selected away. GPT-5.6, which took no part in that selection, answers \textbf{$0.515$} of those
same items with the question hidden, and Sonnet~4.6 $0.320$. Neither figure is a positional
artefact: corrected for each model's own answer-position habit their excess is $+0.265$ and
$+0.081$, the only two excesses in Table~\ref{tab:habit} larger than $0.015$. This is a demonstration
rather than an inference, and it is the paper's central result: filtering a bank against one
model leaves the strongest models in the set reading the options exactly as before.

\paragraph{Position bias explains almost all of the rest.} Unlike the gate, the sweep presents each
item once in its published order, so a model with a positional habit could score above chance
without reading anything. Those habits are large and vendor-specific: Ministral picks
\texttt{A} in $2\%$ of blind trials, Pixtral picks \texttt{B} in $46\%$, Sonnet~4.6 picks
\texttt{D} in $42\%$, and Llama~3.1~8B picks \texttt{A} in $92\%$. So for each model we compute
what its own answer-position distribution alone would earn against the gold-position
distribution of the accepted set, which is \texttt{A}-heavy ($0.290$, $0.245$, $0.241$,
$0.223$). That position-only expectation runs $0.237$--$0.287$ across models, and subtracting it
is what separates recognition from habit.

\begin{table}[H]
\centering\small
\caption{What is left after each model's own positional habit is subtracted. Blind accuracy is
measured on all $8{,}128$ accepted items. ``Habit'' is the
score that model's answer-position distribution alone earns against the gold positions of this
set. The interval is a paired item bootstrap ($10{,}000$ resamples of the $8{,}128$ items, habit
recomputed on each). Only two models have an excess worth the name.}
\label{tab:habit}
\begin{tabular}{lrrrrrrrl}
\toprule
\textbf{Model} & \textbf{A} & \textbf{B} & \textbf{C} & \textbf{D} & \textbf{Blind} &
\textbf{Habit} & \textbf{Excess} & \textbf{95\% CI} \\
\midrule
GPT-5.6       & 0.24 & 0.27 & 0.27 & 0.22 & 0.515 & 0.250 & \textbf{$+0.265$} & [+.254, +.276] \\
Sonnet 4.6    & 0.08 & 0.20 & 0.30 & 0.42 & 0.320 & 0.238 & \textbf{$+0.081$} & [+.072, +.091] \\
DeepSeek R1   & 0.12 & 0.24 & 0.29 & 0.35 & 0.257 & 0.242 & $+0.015$ & [+.005, +.024] \\
Qwen3 32B     & 0.14 & 0.27 & 0.30 & 0.30 & 0.244 & 0.244 & $+0.000$ & [$-$.009, +.009] \\
Ministral 8B  & 0.02 & 0.25 & 0.45 & 0.29 & 0.239 & 0.238 & $+0.001$ & [$-$.008, +.009] \\
Llama 3.3 70B & 0.08 & 0.28 & 0.31 & 0.33 & 0.237 & 0.240 & $-0.003$ & [$-$.012, +.006] \\
Nova Micro    & 0.09 & 0.31 & 0.22 & 0.39 & 0.233 & 0.240 & $-0.007$ & [$-$.016, +.002] \\
Nova Pro      & 0.17 & 0.28 & 0.27 & 0.28 & 0.232 & 0.246 & $-0.014$ & [$-$.023, $-$.005] \\
Gemma 3 12B   & 0.08 & 0.28 & 0.40 & 0.23 & 0.230 & 0.242 & $-0.012$ & [$-$.021, $-$.004] \\
Nova 2 Lite   & 0.03 & 0.30 & 0.29 & 0.39 & 0.225 & 0.237 & $-0.012$ & [$-$.020, $-$.003] \\
Pixtral Large & 0.10 & 0.46 & 0.25 & 0.20 & 0.239 & 0.244 & $-0.005$ & [$-$.014, +.003] \\
\midrule
Llama 3.1 8B  & 0.92 & 0.00 & 0.07 & 0.01 & 0.292 & 0.287 & $+0.005$ & [+.000, +.009] \\
Haiku 4.5     & 0.16 & 0.44 & 0.22 & 0.19 & 0.204 & 0.247 & $-0.043$ & [$-$.052, $-$.035] \\
\bottomrule
\end{tabular}
\end{table}

Two models keep a real excess: GPT-5.6 at $+0.265$ and Sonnet~4.6 at $+0.081$. Every other
held-out model lands within $0.015$ of what its habit alone would earn. DeepSeek~R1 is the one
of them whose interval clears zero, at $+0.015$ $[+0.005, +0.024]$: a detectable trace, an order
smaller than Sonnet's. Rerun through the first-party Anthropic API with raw text kept, Sonnet's
excess reproduces at $+0.083$ and Haiku's at $-0.041$ (Appendix~\ref{app:prompts}). Llama~3.1~8B's $+0.005$ touches zero; the gate's own model
sits $0.043$ \emph{below} its habit, which is what selecting on it was supposed to do. Llama~3.1~8B is the instructive
case: it scores $0.292$ blind, comfortably above chance and above every model we do not call
leaky, purely because it answers \texttt{A} to $92\%$ of items and the key sits at \texttt{A}
in $29\%$ of them. Its excess is $+0.005$. Read without the habit column it would look like a
third leaker; it is not reading the options at all.

\paragraph{The gate found something real, for almost everyone.} On rejected items eleven of the
twelve held-out models score $0.518$--$0.789$, far above chance, and the accepted-versus-rejected
gap is positive for all twelve. The exception is Llama~3.1~8B at $0.337$, which is what a model
that answers \texttt{A} to everything scores on a set where the key is at \texttt{A} more often
than not. For every model that reads the options at all, one model's blind failures are not
idiosyncratic: they identify items that leak to the rest.

\paragraph{The selection also seems to carry a signature of the model that made it.} The gate keeps the items Haiku~4.5 could
\emph{not} answer blind --- items where Haiku's preferred option is a distractor. Any model that
shares Haiku's preferences inherits that preference, and with it the wrong answer. Agreement
with Haiku's blind pick runs $0.42$--$0.53$ for ten of the twelve where independence would give
$0.25$, and it predicts the damage: agreement correlates with the accepted-to-rejected gap at
$r = 0.892$ over twelve models (Table~\ref{tab:imprint}). That correlation leans on one point.
Llama~3.1~8B agrees with Haiku \emph{less} than chance, at $0.182$, because it is not reading
the options, and it has much the smallest gap, $0.045$; drop it and $r$ falls to $0.577$ over
the remaining eleven, which is the honest strength of the relationship among models that do
read. GPT-5.6, the other model outside the band at $0.349$, is also the one the gate damages
least.

\begin{table}[H]
\centering\small
\caption{Agreement with the gating model's blind pick, and the cost of that agreement.
Independence would put agreement at $0.250$. The gap is blind accuracy on rejected items minus
blind accuracy on accepted items; the more a model answers like the gate, the more the gate's
selection costs it.}
\label{tab:imprint}
\begin{tabular}{lrrrr}
\toprule
\textbf{Model} & \textbf{Agrees with Haiku} & \textbf{Accepted} & \textbf{Rejected} & \textbf{Gap} \\
\midrule
Pixtral Large & 0.529 & 0.239 & 0.602 & 0.363 \\
Nova Pro      & 0.504 & 0.232 & 0.645 & 0.413 \\
Llama 3.3 70B & 0.497 & 0.237 & 0.610 & 0.373 \\
Qwen3 32B     & 0.489 & 0.244 & 0.665 & 0.421 \\
Nova 2 Lite   & 0.486 & 0.225 & 0.545 & 0.320 \\
DeepSeek R1   & 0.473 & 0.257 & 0.665 & 0.408 \\
Gemma 3 12B   & 0.462 & 0.230 & 0.595 & 0.365 \\
Ministral 8B  & 0.442 & 0.239 & 0.540 & 0.301 \\
Nova Micro    & 0.430 & 0.233 & 0.518 & 0.285 \\
Sonnet 4.6    & 0.423 & 0.320 & 0.730 & 0.410 \\
GPT-5.6       & 0.349 & 0.515 & 0.789 & 0.274 \\
Llama 3.1 8B  & 0.182 & 0.292 & 0.337 & 0.045 \\
\bottomrule
\end{tabular}
\end{table}

This is the finding that a 400-item sample could not have produced. At that size the nine models
without a real excess all had intervals covering chance and read as ``at chance'' --- a tidy result that
concealed both facts above: that they extract at most $0.015$ once habit is subtracted,
and that what they do extract is shaped by the model that did the selecting.

\paragraph{Leakage scales with capability.} Ranking the eleven held-out models by
full-condition accuracy reproduces their ranking by blind accuracy closely
(Table~\ref{tab:vendors}). Pearson $r = 0.916$ (95\% CI $[0.703, 0.978]$); leaving out any single
model gives $0.838$--$0.943$, and rank correlation is $0.773$.

That correlation is partly mechanical: the $x$-axis is full accuracy on the \emph{same} items as
the $y$-axis, so item-level noise is shared. Replacing the $x$-axis with an independent
capability measure --- full accuracy on the \emph{rejected} items, a disjoint set --- gives
$r = 0.813$ (95\% CI $[0.415, 0.950]$): still positive, with zero outside the interval, and
stable under leave-one-out ($0.809$--$0.854$). Note what the association now rests on: with the
blind axis compressed into a band of $0.032$ for nine of the eleven, it is carried mostly by
the two models that leak. Among the nine alone, on the same disjoint axis, it is $r = 0.713$ ---
suggestive, but over a band narrower than the intervals of the quantity it is meant to explain.
We report an observed association.

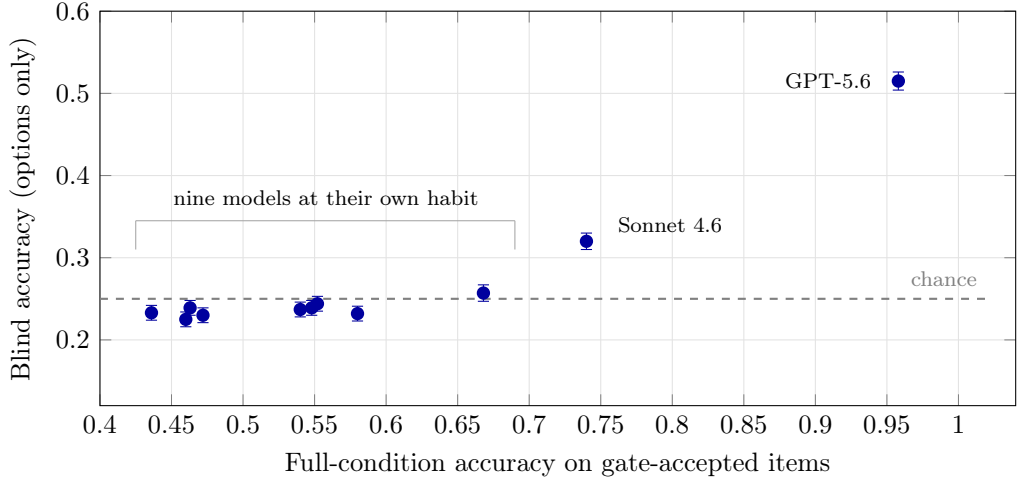
\begin{figure}[H]
\centering
\begin{tikzpicture}
\begin{axis}[
  width=0.86\textwidth, height=6.8cm,
  xlabel={Full-condition accuracy on gate-accepted items},
  ylabel={Blind accuracy (options only)},
  xmin=0.40, xmax=1.04, ymin=0.12, ymax=0.60,
  grid=major, grid style={gray!22},
  tick label style={font=\small}, label style={font=\small},
  every axis plot/.append style={thick},
]
\addplot[domain=0.40:1.02, dashed, gray] {0.25};
\node[gray, font=\scriptsize, anchor=south east] at (axis cs:1.02,0.253) {chance};
\addplot[only marks, mark=*, mark size=2.1pt, color=blue!62!black,
         error bars/.cd, y dir=both, y explicit]
  coordinates {
    (0.436, 0.233) +- (0, 0.009)
    (0.460, 0.225) +- (0, 0.009)
    (0.463, 0.239) +- (0, 0.009)
    (0.472, 0.230) +- (0, 0.009)
    (0.540, 0.237) +- (0, 0.009)
    (0.548, 0.239) +- (0, 0.009)
    (0.552, 0.244) +- (0, 0.009)
    (0.580, 0.232) +- (0, 0.009)
    (0.668, 0.257) +- (0, 0.010)
    (0.740, 0.320) +- (0, 0.010)
    (0.958, 0.515) +- (0, 0.011)
  };
\node[font=\scriptsize, anchor=south] at (axis cs:0.558,0.352) {nine models at their own habit};
\draw[gray!70] (axis cs:0.425,0.345) -- (axis cs:0.425,0.310);
\draw[gray!70] (axis cs:0.690,0.345) -- (axis cs:0.690,0.310);
\draw[gray!70] (axis cs:0.425,0.345) -- (axis cs:0.690,0.345);
\node[font=\scriptsize, anchor=west] at (axis cs:0.755,0.340) {Sonnet 4.6};
\node[font=\scriptsize, anchor=east] at (axis cs:0.946,0.515) {GPT-5.6};
\end{axis}
\end{tikzpicture}
\caption{Blind accuracy against full-condition accuracy on the gate-accepted items, eleven
held-out models from eight vendors. Bars are 95\% Wilson intervals on the blind estimate, now
measured on the whole accepted set rather than a 400-item sample, which is why they are short;
the horizontal position still carries the sample's error. The plotted correlation,
$r = 0.916$, shares items between the two axes; against an independent capability estimate on
disjoint items it is $r = 0.813$ (\S\ref{sec:vendors}). The gate was built with a model absent
from this plot, and Llama~3.1~8B is absent because it answers \texttt{A} to $92\%$ of items. The
bracketed nine sit within $0.015$ of their own positional habit (Table~\ref{tab:habit}).}
\label{fig:scaling}
\end{figure}

The more robust statement is a ratio, and Figure~\ref{fig:scaling} shows its shape. Blind
accuracy is a fixed fraction of full accuracy across the capability range: $0.384$--$0.538$,
mean $0.463$. That arithmetic holds, but it should not be read as ``a model recovers $46\%$ of
its competence without the question''. For nine of the eleven the blind score is their
positional habit and nothing else, so the fraction describes where habit happens to land, not
partial competence. The reading survives only for the two models with a real excess.

The implication is the paper's main point, and it comes in two parts that should not be
conflated. The first we observe directly: a bank filtered until one model is at chance is not
clean for a stronger one. Our accepted set was gated against Haiku~4.5, and GPT-5.6 answers
$0.515$ of it blind. That is a demonstration, not an inference.

The second is a prediction. If blind accuracy keeps tracking capability beyond the range we
tested, then filtering against the models available today will not hold against the models
available later, and each round of filtering will remove the items that discriminate best.
We have eleven models spanning full-condition accuracy from $0.436$ to $0.958$; whether the
relationship continues above that is untested, and we state it as a conjecture rather than a
result.

\subsection{Five more models, and whether the excess is recall}
\label{sec:crossvendor}

In November 2026 we scored five further models on the same $8{,}128$ accepted items, through
Azure AI Foundry and Google Vertex AI rather than Bedrock, with the same prompt, option order,
temperature~0 and $4{,}096$-token budget, keeping every reply (Appendix~\ref{app:models}). All five were released before our own release
of the bank on 15~August 2026. Table~\ref{tab:crossvendor}
is computed exactly as Table~\ref{tab:habit}.

\begin{table}[H]
\centering\small
\caption{Five further models and one control on the $8{,}128$ accepted items, November 2026,
estimator and bootstrap as in Table~\ref{tab:habit}. Gemini~3.1~Pro ran at
\texttt{thinkingLevel} \textsc{low}; at its default thinking it scores $0.060$
$[+0.016, +0.104]$ higher on the 385 accepted items of the 600-item sample it completed within
budget, so its row is a lower bound. GPT-4o (\texttt{gpt-4o-2024-11-20}) is the
contamination control: its training data are reported to end in October 2023, before the bank
was published, though the snapshot itself is from November 2024.}
\label{tab:crossvendor}
\begin{tabular}{lrrrrrrrl}
\toprule
\textbf{Model} & \textbf{A} & \textbf{B} & \textbf{C} & \textbf{D} & \textbf{Blind} &
\textbf{Habit} & \textbf{Excess} & \textbf{95\% CI} \\
\midrule
Gemini 3.1 Pro   & 0.28 & 0.26 & 0.26 & 0.20 & 0.493 & 0.253 & \textbf{$+0.240$} & [+.229, +.251] \\
GPT-5.4          & 0.13 & 0.32 & 0.29 & 0.26 & 0.349 & 0.244 & \textbf{$+0.105$} & [+.095, +.115] \\
DeepSeek V4 Pro  & 0.02 & 0.59 & 0.29 & 0.10 & 0.274 & 0.243 & $+0.031$ & [+.023, +.039] \\
gpt-oss-120b     & 0.26 & 0.24 & 0.26 & 0.24 & 0.267 & 0.251 & $+0.016$ & [+.006, +.026] \\
Mistral Large 3  & 0.05 & 0.21 & 0.31 & 0.43 & 0.253 & 0.237 & $+0.016$ & [+.007, +.025] \\
\midrule
GPT-4o (control) & 0.22 & 0.27 & 0.35 & 0.16 & 0.263 & 0.250 & $+0.012$ & [+.003, +.022] \\
\bottomrule
\end{tabular}
\end{table}

The pattern of Table~\ref{tab:habit} recurs across vendors. Two frontier models keep a large
excess, Gemini~3.1~Pro at $+0.240$ and GPT-5.4 at $+0.105$; the other three sit at $+0.031$ or
less, with intervals clear of zero but an order smaller. A gate built with Haiku~4.5 leaves
Google's model almost as much to extract as OpenAI's, so the failure to transfer is not a
property of one vendor pair.

\paragraph{Recall of a public bank does not explain the excess.} The bank and its keys have
been public since 15~July 2024 (\S\ref{sec:corpus}), and every strong model we score was trained
after that, so blind success could be recall of the item rather than inference from its options.
We checked this in two ways. The first is a control whose reported training cutoff precedes publication: GPT-4o, at
$+0.012$ $[+0.003, +0.022]$ (Table~\ref{tab:crossvendor}). It is also weaker than the two models
that extract, so on its own it cannot separate capability from exposure. The second is a direct
probe on the 600-item sample. We hide one \emph{distractor} and ask the model to reproduce it
verbatim, in the manner of testset slot guessing \citep{deng2024investigating}: a distractor's
wording is the examiner's invention and cannot be derived from the law. On items with a
positive stem and a hidden option of at least 20 characters ($n = 406$), Gemini~3.1~Pro
reproduces it near-verbatim (word-sequence similarity $\geq 0.9$) in $8.6\%$ of cases, GPT-5.4 in
$6.4\%$ and GPT-4o in $1.0\%$, though GPT-4o refuses $81\%$ of these requests and reaches $5.1\%$
on those it answers. Some exposure is plausible at that rate. It does not, however, carry over
to the blind condition: on the 309 accepted items of the sample with a hidden option of that
length, the items whose hidden distractor a model reproduces are answered blind no more often
than the rest ($-0.033$ $[-0.20, +0.14]$ for Gemini~3.1~Pro, $-0.053$ $[-0.24, +0.14]$ for
GPT-5.4; $-0.160$ and $+0.013$ on positive stems alone, both intervals spanning zero), and the
correlation between similarity and blind correctness is $-0.05$ and $0.00$. Asked instead to state the keyed answer from the
question alone, the three models reproduce it near-verbatim in $19.4\%$, $7.8\%$ and $5.2\%$ of
items. That score correlates weakly and positively with blind success in all three
($r = 0.08$, $0.11$, $0.18$), and only for GPT-4o, whose reported cutoff precedes the bank, does the
interval clear zero ($[+0.06, +0.31]$): the pattern knowing the law rather than the bank would produce.
We find no evidence that the excess is recall; a model with frontier capability and a training
cutoff before July 2024 would settle the question, and none was available to us.

\subsection{The numbers are not an artefact of how we asked}
\label{sec:ablation}

Every blind figure above comes from one prompt, so we re-ran the blind condition on the 400
gate-accepted items under five phrasings: the original; a minimal form with no framing at all
(``choose one option''); the original with the word \emph{legal} removed, to test whether naming
the domain is itself what primes recognition; an English instruction over Ukrainian options; and
an explicit ``this is a guessing task'' framing. Two leaking models and one that sits at chance
were run on all five.

\begin{table}[H]
\centering\small
\caption{Blind accuracy under five prompt phrasings, 400 gate-accepted items. Chance is
$0.250$. These figures are on a 400-item subset of the accepted set, not the whole of it, so
they sit above the full-scale numbers in Table~\ref{tab:vendors} --- for these three models the
subset runs $0.009$ to $0.024$ leakier than the set it was drawn from, and every model in it
inherits that. What
the table is for is the \emph{spread} within a model, which the shared sample does not
distort.}
\label{tab:ablation}
\begin{tabular}{lrrrr}
\toprule
\textbf{Prompt} & \textbf{Sonnet 4.6} & \textbf{GPT-5.6} & \textbf{Nova Pro} \\
\midrule
Original                      & 0.343 & 0.524 & 0.255 \\
Minimal, no framing           & 0.323 & 0.463 & 0.265 \\
Domain word removed           & 0.310 & 0.490 & 0.236 \\
English instruction           & 0.325 & 0.521 & 0.240 \\
Framed as guessing            & 0.330 & 0.479 & 0.220 \\
\midrule
Spread                        & 0.033 & 0.061 & 0.045 \\
\bottomrule
\end{tabular}
\end{table}

Three things follow. The leak survives every phrasing: Sonnet~4.6's worst variant is $0.310$
with a lower confidence bound of $0.267$, still above chance, and GPT-5.6's worst is $0.463$.
Naming the domain helps a little and explains almost nothing --- dropping \emph{legal} costs
$0.033$ for Sonnet and $0.034$ for GPT-5.6, leaving both far above chance. And the control
behaves: Nova Pro stays at chance under all five phrasings, every interval covering $0.250$, so
the probe does not manufacture signal where there is none. The within-model spread,
$0.033$--$0.061$, is far smaller than the $0.18$ that separates Sonnet from GPT-5.6 on these
items, so which models leak is not a function of how we asked.

That covers the blind prompt, which leaves three other things fixed at one realisation each,
all of them underneath the ratio: the wording of the \emph{full} condition, which is its
denominator; the labels on the options, which are Latin \texttt{A)}--\texttt{D)} set over
Ukrainian text, so that a letter prior would be indistinguishable from recognition of content;
and the order of the options. We vary the first two here and the third in
\S\ref{sec:position}, moving one factor at a time from the configuration used throughout.

\begin{table}[H]
\centering\small
\caption{Presentation robustness on the same 400 gate-accepted items. Blind chance is $0.250$.
Marker rows change only the labels; the Latin rows repeat the baseline for reference. No model
answered with a Latin letter when shown Cyrillic labels, so the conventions are not being
conflated.}
\label{tab:presentation}
\begin{tabular}{llrrr}
\toprule
& & \textbf{Sonnet 4.6} & \textbf{GPT-5.6} & \textbf{Nova Pro} \\
\midrule
\multicolumn{5}{@{}l}{\emph{Full-condition wording}} \\
& Original                  & 0.738 & 0.957 & 0.578 \\
& Minimal, no framing       & 0.740 & 0.963 & 0.610 \\
& Exam framing removed      & 0.755 & 0.962 & 0.573 \\
& English instruction       & 0.750 & 0.962 & 0.578 \\
& Expert persona            & 0.757 & 0.960 & 0.598 \\
& \emph{spread}             & 0.020 & 0.005 & 0.037 \\
\midrule
\multicolumn{5}{@{}l}{\emph{Option labels, blind}} \\
& Latin \texttt{A)}--\texttt{D)}   & 0.343 & 0.524 & 0.255 \\
& Cyrillic \foreignlanguage{ukrainian}{\texttt{А)}--\texttt{Г)}} & 0.360 & 0.519 & 0.273 \\
& Numeric \texttt{1)}--\texttt{4)} & 0.347 & 0.507 & 0.263 \\
\midrule
\multicolumn{5}{@{}l}{\emph{Option labels, full}} \\
& Latin \texttt{A)}--\texttt{D)}   & 0.738 & 0.957 & 0.578 \\
& Cyrillic \foreignlanguage{ukrainian}{\texttt{А)}--\texttt{Г)}} & 0.750 & 0.962 & 0.583 \\
& Numeric \texttt{1)}--\texttt{4)} & 0.765 & 0.967 & 0.570 \\
\bottomrule
\end{tabular}
\end{table}

Neither matters. The full condition moves by $0.005$--$0.037$ across five phrasings, so the
denominator of the ratio is not a property of our prompt: recomputed against each of them, the
blind-to-full ratio spans $0.452$--$0.464$ for Sonnet and $0.544$--$0.547$ for GPT-5.6, a spread
far narrower than the $0.384$--$0.538$ range we report across models. And the labels are inert in both
conditions and for all three models, the largest shift being $0.027$. The Latin alphabet was
not doing the work.

\subsection{Position bias is a different thing from the leak}
\label{sec:position}

The blind numbers so far use the option order as printed in the bank. A model that simply
favours one slot would score above chance whenever the key happened to be there, which in a
single-order measurement is indistinguishable from recognising the option's content. So we
forced the key into each of the four slots in turn, keeping the distractors in their relative
order: 400 items $\times$ 4 placements $\times$ 3 models, blind.

This yields a test that needs no modelling assumption. If a model's choice were independent of
what the options say, then $P(\text{correct} \mid \text{key in slot } j)$ is just $q_j$, the
rate at which it reaches for slot $j$; and because the $q_j$ sum to one, the four accuracies
must sum to one as well. Their mean is then exactly chance, $0.250$, however lopsided the
preference. Whatever exceeds that is content. Since each item contributes four correlated
observations, we take the interval at item level rather than treating $1{,}600$ responses as
independent.

\begin{figure}[H]
\centering
% Created by tikzDevice version 0.12.6 on 2026-08-15 15:28:26
% !TEX encoding = UTF-8 Unicode
\begin{tikzpicture}[x=1pt,y=1pt]
\definecolor{fillColor}{RGB}{255,255,255}
\path[use as bounding box,fill=fillColor,fill opacity=0.00] (0,0) rectangle (404.71,158.99);
\begin{scope}
\path[clip] (  0.00,  0.00) rectangle (404.71,158.99);
\definecolor{fillColor}{RGB}{255,255,255}

\path[fill=fillColor] (  0.00,  0.00) rectangle (404.71,158.99);
\end{scope}
\begin{scope}
\path[clip] ( 25.45, 22.61) rectangle (147.53,142.73);
\definecolor{drawColor}{gray}{0.92}

\path[draw=drawColor,line width= 0.5pt,line join=round] ( 25.45, 22.61) --
	(147.53, 22.61);

\path[draw=drawColor,line width= 0.5pt,line join=round] ( 25.45, 56.25) --
	(147.53, 56.25);

\path[draw=drawColor,line width= 0.5pt,line join=round] ( 25.45, 89.90) --
	(147.53, 89.90);

\path[draw=drawColor,line width= 0.5pt,line join=round] ( 25.45,123.55) --
	(147.53,123.55);
\definecolor{fillColor}{RGB}{44,93,143}

\path[fill=fillColor,fill opacity=0.90] ( 33.88, 22.61) rectangle ( 51.90, 50.79);

\path[fill=fillColor,fill opacity=0.90] ( 62.94, 22.61) rectangle ( 80.97, 68.87);

\path[fill=fillColor,fill opacity=0.90] ( 92.01, 22.61) rectangle (110.04, 72.24);

\path[fill=fillColor,fill opacity=0.90] (121.08, 22.61) rectangle (139.10, 81.07);
\definecolor{drawColor}{RGB}{179,118,42}
\definecolor{fillColor}{RGB}{179,118,42}

\path[draw=drawColor,line width= 0.3pt,line join=round,line cap=round,fill=fillColor] ( 42.89, 50.26) circle (  1.90);

\path[draw=drawColor,line width= 0.3pt,line join=round,line cap=round,fill=fillColor] ( 71.96, 65.29) circle (  1.90);

\path[draw=drawColor,line width= 0.3pt,line join=round,line cap=round,fill=fillColor] (101.02, 69.07) circle (  1.90);

\path[draw=drawColor,line width= 0.3pt,line join=round,line cap=round,fill=fillColor] (130.09, 74.02) circle (  1.90);

\path[draw=drawColor,line width= 0.5pt,line join=round] ( 42.89, 50.26) --
	( 71.96, 65.29) --
	(101.02, 69.07) --
	(130.09, 74.02);
\end{scope}
\begin{scope}
\path[clip] (152.03, 22.61) rectangle (274.12,142.73);
\definecolor{drawColor}{gray}{0.92}

\path[draw=drawColor,line width= 0.5pt,line join=round] (152.03, 22.61) --
	(274.12, 22.61);

\path[draw=drawColor,line width= 0.5pt,line join=round] (152.03, 56.25) --
	(274.12, 56.25);

\path[draw=drawColor,line width= 0.5pt,line join=round] (152.03, 89.90) --
	(274.12, 89.90);

\path[draw=drawColor,line width= 0.5pt,line join=round] (152.03,123.55) --
	(274.12,123.55);
\definecolor{fillColor}{RGB}{44,93,143}

\path[fill=fillColor,fill opacity=0.90] (160.46, 22.61) rectangle (178.49, 43.22);

\path[fill=fillColor,fill opacity=0.90] (189.53, 22.61) rectangle (207.56, 66.35);

\path[fill=fillColor,fill opacity=0.90] (218.60, 22.61) rectangle (236.62, 89.06);

\path[fill=fillColor,fill opacity=0.90] (247.67, 22.61) rectangle (265.69,131.96);
\definecolor{drawColor}{RGB}{179,118,42}
\definecolor{fillColor}{RGB}{179,118,42}

\path[draw=drawColor,line width= 0.3pt,line join=round,line cap=round,fill=fillColor] (169.48, 35.54) circle (  1.90);

\path[draw=drawColor,line width= 0.3pt,line join=round,line cap=round,fill=fillColor] (198.54, 53.51) circle (  1.90);

\path[draw=drawColor,line width= 0.3pt,line join=round,line cap=round,fill=fillColor] (227.61, 71.29) circle (  1.90);

\path[draw=drawColor,line width= 0.3pt,line join=round,line cap=round,fill=fillColor] (256.68, 98.31) circle (  1.90);

\path[draw=drawColor,line width= 0.5pt,line join=round] (169.48, 35.54) --
	(198.54, 53.51) --
	(227.61, 71.29) --
	(256.68, 98.31);
\end{scope}
\begin{scope}
\path[clip] (278.62, 22.61) rectangle (400.71,142.73);
\definecolor{drawColor}{gray}{0.92}

\path[draw=drawColor,line width= 0.5pt,line join=round] (278.62, 22.61) --
	(400.71, 22.61);

\path[draw=drawColor,line width= 0.5pt,line join=round] (278.62, 56.25) --
	(400.71, 56.25);

\path[draw=drawColor,line width= 0.5pt,line join=round] (278.62, 89.90) --
	(400.71, 89.90);

\path[draw=drawColor,line width= 0.5pt,line join=round] (278.62,123.55) --
	(400.71,123.55);
\definecolor{fillColor}{RGB}{44,93,143}

\path[fill=fillColor,fill opacity=0.90] (287.05, 22.61) rectangle (305.08,100.84);

\path[fill=fillColor,fill opacity=0.90] (316.12, 22.61) rectangle (334.14,115.14);

\path[fill=fillColor,fill opacity=0.90] (345.19, 22.61) rectangle (363.21,107.15);

\path[fill=fillColor,fill opacity=0.90] (374.26, 22.61) rectangle (392.28,105.25);
\definecolor{drawColor}{RGB}{179,118,42}
\definecolor{fillColor}{RGB}{179,118,42}

\path[draw=drawColor,line width= 0.3pt,line join=round,line cap=round,fill=fillColor] (296.06, 62.90) circle (  1.90);

\path[draw=drawColor,line width= 0.3pt,line join=round,line cap=round,fill=fillColor] (325.13, 67.85) circle (  1.90);

\path[draw=drawColor,line width= 0.3pt,line join=round,line cap=round,fill=fillColor] (354.20, 65.32) circle (  1.90);

\path[draw=drawColor,line width= 0.3pt,line join=round,line cap=round,fill=fillColor] (383.27, 62.58) circle (  1.90);

\path[draw=drawColor,line width= 0.5pt,line join=round] (296.06, 62.90) --
	(325.13, 67.85) --
	(354.20, 65.32) --
	(383.27, 62.58);
\end{scope}
\begin{scope}
\path[clip] ( 25.45,142.73) rectangle (147.53,156.99);
\definecolor{drawColor}{gray}{0.10}

\node[text=drawColor,anchor=base,inner sep=0pt, outer sep=0pt, scale=  0.80] at ( 86.49,147.11) {Nova Pro (sum 1.08)};
\end{scope}
\begin{scope}
\path[clip] (152.03,142.73) rectangle (274.12,156.99);
\definecolor{drawColor}{gray}{0.10}

\node[text=drawColor,anchor=base,inner sep=0pt, outer sep=0pt, scale=  0.80] at (213.08,147.11) {Sonnet 4.6 (sum 1.43)};
\end{scope}
\begin{scope}
\path[clip] (278.62,142.73) rectangle (400.71,156.99);
\definecolor{drawColor}{gray}{0.10}

\node[text=drawColor,anchor=base,inner sep=0pt, outer sep=0pt, scale=  0.80] at (339.67,147.11) {GPT-5.6 (sum 2.01)};
\end{scope}
\begin{scope}
\path[clip] (  0.00,  0.00) rectangle (404.71,158.99);
\definecolor{drawColor}{gray}{0.30}

\node[text=drawColor,anchor=base,inner sep=0pt, outer sep=0pt, scale=  0.72] at ( 42.89, 13.60) {A};

\node[text=drawColor,anchor=base,inner sep=0pt, outer sep=0pt, scale=  0.72] at ( 71.96, 13.60) {B};

\node[text=drawColor,anchor=base,inner sep=0pt, outer sep=0pt, scale=  0.72] at (101.02, 13.60) {C};

\node[text=drawColor,anchor=base,inner sep=0pt, outer sep=0pt, scale=  0.72] at (130.09, 13.60) {D};
\end{scope}
\begin{scope}
\path[clip] (  0.00,  0.00) rectangle (404.71,158.99);
\definecolor{drawColor}{gray}{0.30}

\node[text=drawColor,anchor=base,inner sep=0pt, outer sep=0pt, scale=  0.72] at (169.48, 13.60) {A};

\node[text=drawColor,anchor=base,inner sep=0pt, outer sep=0pt, scale=  0.72] at (198.54, 13.60) {B};

\node[text=drawColor,anchor=base,inner sep=0pt, outer sep=0pt, scale=  0.72] at (227.61, 13.60) {C};

\node[text=drawColor,anchor=base,inner sep=0pt, outer sep=0pt, scale=  0.72] at (256.68, 13.60) {D};
\end{scope}
\begin{scope}
\path[clip] (  0.00,  0.00) rectangle (404.71,158.99);
\definecolor{drawColor}{gray}{0.30}

\node[text=drawColor,anchor=base,inner sep=0pt, outer sep=0pt, scale=  0.72] at (296.06, 13.60) {A};

\node[text=drawColor,anchor=base,inner sep=0pt, outer sep=0pt, scale=  0.72] at (325.13, 13.60) {B};

\node[text=drawColor,anchor=base,inner sep=0pt, outer sep=0pt, scale=  0.72] at (354.20, 13.60) {C};

\node[text=drawColor,anchor=base,inner sep=0pt, outer sep=0pt, scale=  0.72] at (383.27, 13.60) {D};
\end{scope}
\begin{scope}
\path[clip] (  0.00,  0.00) rectangle (404.71,158.99);
\definecolor{drawColor}{gray}{0.30}

\node[text=drawColor,anchor=base east,inner sep=0pt, outer sep=0pt, scale=  0.72] at ( 21.40, 20.13) {0.0};

\node[text=drawColor,anchor=base east,inner sep=0pt, outer sep=0pt, scale=  0.72] at ( 21.40, 53.77) {0.2};

\node[text=drawColor,anchor=base east,inner sep=0pt, outer sep=0pt, scale=  0.72] at ( 21.40, 87.42) {0.4};

\node[text=drawColor,anchor=base east,inner sep=0pt, outer sep=0pt, scale=  0.72] at ( 21.40,121.07) {0.6};
\end{scope}
\begin{scope}
\path[clip] (  0.00,  0.00) rectangle (404.71,158.99);
\definecolor{drawColor}{RGB}{0,0,0}

\node[text=drawColor,anchor=base,inner sep=0pt, outer sep=0pt, scale=  0.90] at (213.08,  3.75) {Slot holding the key};
\end{scope}
\begin{scope}
\path[clip] (  0.00,  0.00) rectangle (404.71,158.99);
\definecolor{drawColor}{RGB}{0,0,0}

\node[text=drawColor,rotate= 90.00,anchor=base,inner sep=0pt, outer sep=0pt, scale=  0.90] at (  8.20, 82.67) {Blind accuracy};
\end{scope}
\end{tikzpicture}
\caption{Blind accuracy with the key forced into each slot (bars) against how often the model
picks that slot at all (points). Under content-independence the two coincide and the bars sum
to $1.00$. Nova Pro has a real slot preference and no gap; Sonnet has the largest preference in
the set and a clear gap; GPT-5.6 has almost no preference and the largest gap of all.}
\label{fig:position}
\end{figure}
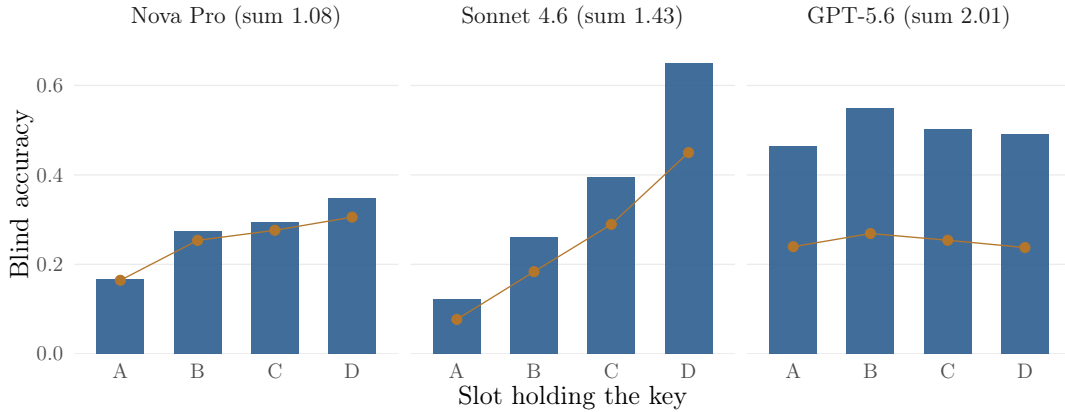

Figure~\ref{fig:position} shows the three cases and Table~\ref{tab:position} gives the numbers.
Sonnet~4.6 turns out to carry a severe positional prior --- it picks the last slot $0.450$ of
the time and the first $0.077$ --- which drags its blind accuracy from $0.122$ with the key at
\texttt{A} to $0.650$ with the key at \texttt{D}, a spread of $0.528$, an order of magnitude
larger than anything in Table~\ref{tab:presentation}. But the accuracies sum to $1.427$, not
$1.000$, and the lift over the preference is positive in every slot, so content is being read
on top of the prior. Nova Pro, the control, has a genuine preference of its own and no content:
it sums to $1.085$ and its position-free estimate, $0.271$, does not separate from chance.
GPT-5.6 is the opposite of Sonnet, picking the four slots almost uniformly and scoring near
$0.50$ wherever the key sits.

\begin{table}[H]
\centering\small
\caption{Blind solvability with position bias removed by construction: the mean over the four
forced placements, with an item-level $95\%$ interval on 400 items. The single-order column is
the same 400-item measurement, not the full-scale figure of Table~\ref{tab:vendors}.}
\label{tab:position}
\begin{tabular}{lrrrrr}
\toprule
\textbf{Model} & \textbf{Slot spread} & \textbf{Sum} & \textbf{Position-free} & \textbf{95\% CI}
& \textbf{Single-order} \\
\midrule
Nova Pro   & 0.180 & 1.085 & 0.271 & [0.239, 0.304] & 0.255 \\
Sonnet 4.6 & 0.528 & 1.427 & 0.357 & [0.325, 0.389] & 0.343 \\
GPT-5.6    & 0.085 & 2.009 & 0.502 & [0.461, 0.542] & 0.524 \\
\bottomrule
\end{tabular}
\end{table}

Two things follow. Order does not carry the result: on this subset the position-free estimates
sit within $0.022$ of the single-order values measured on the same items, and for Sonnet the
single-order figure is if anything the conservative one, so nothing here depends on where the
key happened to be printed. And position bias and leak exploitation are orthogonal. The model with the largest
slot preference is not the one that leaks most, and the model with almost none leaks most of
all. This matters for remedies: permuting options, the standard fix for position bias and the
one our own gate relies on, does nothing whatever about content that is recognisable on its own.

The labelling sweep of \S\ref{sec:ablation} also speaks to the mechanism.
\citet{zheng2024robust} trace selection bias to token bias --- probability mass attached to the
option-ID tokens themselves --- which predicts that the preference should move when the tokens
change. It largely does not. Sonnet~4.6 picks the first slot $0.092$ of the time under Latin
\texttt{A}--\texttt{D}, $0.098$ under Cyrillic
\foreignlanguage{ukrainian}{\texttt{А}--\texttt{Г}} and $0.090$ under digits \texttt{1}--\texttt{4};
Nova Pro likewise stays in a narrow band ($0.180$, $0.147$, $0.182$). What the labels do change is
the size of the pull towards the last slot, which for Sonnet runs $0.415$, $0.477$ and $0.310$.
On our items the bias is positional first and token-dependent second.

\subsection{Reasoning at inference time is not what makes options solvable}
\label{sec:reasoning}

Our models run at vendor defaults, so the set mixes models that reason before answering with
models that do not --- as \citet{fan2026lexam} also do, and for the same reason: reasoning
cannot be switched off in GPT-5.6, which rejects \texttt{reasoning\_effort} on Bedrock as it
rejects \texttt{temperature}. That leaves an objection open: perhaps the $0.18$ separating
GPT-5.6 from Sonnet blind is inference-time compute rather than what they read from the options.
What cannot be equalised downwards can be moved upwards and measured.

\begin{table}[H]
\centering\small
\caption{Reasoning interventions on the 400 gate-accepted items. $p$-values are McNemar tests
on the paired items against that model's direct-answer row. The extended-thinking row is
reported for completeness only: Anthropic's API refuses temperature~0 when thinking is enabled,
so that row moves two things at once.}
\label{tab:reasoning}
\begin{tabular}{llrrrr}
\toprule
\textbf{Model} & \textbf{Condition} & \textbf{Blind} & \textbf{$p$} & \textbf{Full} & \textbf{Out tok.} \\
\midrule
Sonnet 4.6  & direct answer        & 0.343 & ---   & 0.738 & 4 \\
Sonnet 4.6  & extended thinking on & 0.343 & ---   & 0.785 & 59 \\
Sonnet 4.6  & step-by-step prompt  & 0.388 & 0.139 & 0.805 & 619 \\
\midrule
Nova Pro    & direct answer        & 0.255 & ---   & 0.578 & 260 \\
Nova Pro    & step-by-step prompt  & 0.228 & 0.284 & 0.578 & 530 \\
\midrule
DeepSeek R1 & 2{,}048-token budget & 0.253 & ---   & 0.668 & --- \\
DeepSeek R1 & 4{,}096-token budget & 0.231 & ---   & 0.648 & 1{,}363 \\
\bottomrule
\end{tabular}
\end{table}

Four measurements agree. Enabling extended thinking leaves Sonnet's blind accuracy exactly where
it was, though that alone proves little --- the model declines the budget, spending 59 output
tokens of an allowed 2{,}000. Forcing the reasoning at the prompt, as LEXam does for every
model, makes it reason at 619 tokens against 4, and moves blind accuracy from $0.343$ to
$0.388$, which a paired test does not separate from noise. The same instruction on the same
items raises the \emph{full} condition from $0.738$ to $0.805$ ($p = 0.001$). Reasoning that
demonstrably helps when the question is present does not measurably help when it is absent,
which is what one should expect: with no question there is nothing to reason towards. Nor does
it manufacture the ability --- Nova Pro under the same instruction stays at chance.

Across models the association fails in both directions: DeepSeek~R1 spends $1{,}363$ output
tokens per blind item and scores $0.231$, at chance, while GPT-5.6 also reasons and scores
$0.524$ on this subset. The two reasoning models sit at the two extremes, with a conventional
model between them. Inference-time reasoning is neither necessary nor sufficient for
option-only solvability.

Re-running R1 at LEXam's 4{,}096-token budget rather than ours also resolves the one real
parsing casualty in this paper: on the blind condition over accepted items, unparsed responses
fall from $18.0\%$ to $1.5\%$ --- the $13.1\%$ quoted elsewhere pools all four of its cells ---
while the estimates move only within their intervals ($0.253 \to 0.231$ blind on $328$ then
$394$ parsed items, $0.668 \to 0.648$ full). The truncation cost
coverage; it was not biasing the numbers.

\section{A Benchmark That Does Not Leak, and Why}
\label{sec:lexam}

Filtering failed, and rewriting distractors failed. The one thing that works is not a repair at
all but a property some banks have from the start. If option-only solvability were a general
property of legal exam items, it should appear in other exam-derived benchmarks. We ran the same blind probe on the four-choice split of LEXam
\citep{fan2026lexam}, 1{,}655 items from Swiss university law exams, with the two models that
leaked most on our bank.

\begin{table}[H]
\centering
\caption{The same blind probe on LEXam's four-choice split. Chance is $0.250$. $n$ counts
responses from which a letter could be parsed; GPT-5.6 produced 34 unparseable answers in
the full condition and none in the blind one.}
\label{tab:lexam}
\begin{tabular}{llrrr}
\toprule
\textbf{Model} & \textbf{Condition} & \textbf{$n$} & \textbf{Accuracy} & \textbf{95\% CI} \\
\midrule
Sonnet 4.6  & blind & 1655 & 0.228 & [0.208, 0.249] \\
Sonnet 4.6  & full  & 1655 & 0.642 & [0.619, 0.665] \\
GPT-5.6     & blind & 1655 & 0.228 & [0.208, 0.249] \\
GPT-5.6     & full  & 1621 & 0.808 & [0.788, 0.827] \\
\bottomrule
\end{tabular}
\end{table}

Both models sit at chance, with the upper confidence bound below $0.250$. The same GPT-5.6 that
recovers $0.515$ of our gate-accepted items blind recovers nothing here.

\begin{figure}[H]
\centering
\input{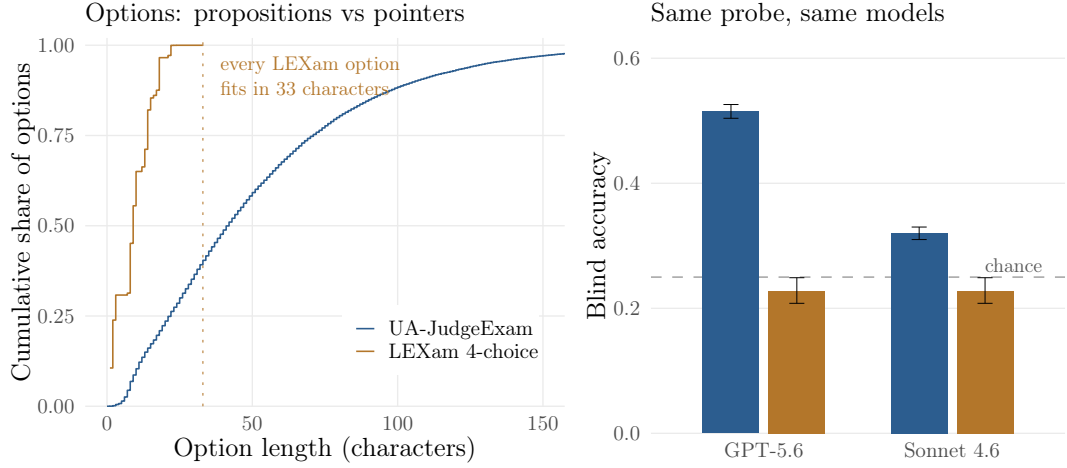}
\caption{Left: the cumulative distribution of option length in the two banks. LEXam's curve is
complete at 33 characters --- its longest option anywhere --- while $59.6\%$ of our options are
longer than that. Ours are legal propositions that can be judged on their own; LEXam's are
pointers into a list given in the stem. Right: what the same two models score with the question
hidden. The design choice on the left produces the difference on the right.}
\label{fig:format}
\end{figure}

The reason is visible in the items (Figure~\ref{fig:format}). In \emph{every one} of LEXam's
$1{,}655$ four-choice items, all four options are \emph{references} to statements enumerated in
the stem --- ``i und iii'', ``ii and iii'', ``none of the statements''. The median option is 9
characters against 42 in our bank, and the point is sharper than the median: the longest option
anywhere in the split is 33 characters, \foreignlanguage{ngerman}{``ii, iii, iv, v, vii, viii,
und ix''}. No legal proposition of any kind appears in an option position. A reference carries
no content of its own, so there is nothing for a reader to recognise when the stem is hidden.
LEXam is immune to this failure by construction, not by filtering.

This cuts two ways. It validates the probe: an instrument that reports leakage should report
chance where chance is the ground truth, and it does. And it relocates our finding. Option-only
solvability is not a property of legal exams, nor of any particular jurisdiction or language.
It is a property of \emph{item format}: options that are self-contained legal propositions can
be judged on their own, and options that are pointers cannot. Banks written in the first style
--- which includes most professional licensing material we are aware of, and ours --- need the
blind baseline reported. Banks written in the second do not.

\section{Remedies: One Clear Failure, Two Cautions}
\label{sec:failed}

Filtering is not the only repair one might try, so we tried the others. None of them lands on
chance either, which is the same failure in a different guise: an item set that a model gets
wrong at a predictable rate is not a neutral item set.

\paragraph{Swapping in real answers overshoots.} If distractors leak because they read as
implausible law, replace them with propositions that are real law: the correct answers of other
items, chosen from the same specialisation, matched on answer type and length, with the key's
position re-randomised. Blind accuracy for Sonnet~4.6 moves from $0.386$ $[0.321, 0.455]$ to
$\mathbf{0.168}$ $[0.122, 0.226]$ ($n=197$, intervals disjoint) while full accuracy is
essentially preserved ($0.792 \to 0.746$). The mechanism is
real and controllable --- and the result is useless. A set on which a model scores $0.168$
blind is as exploitable as one on which it scores $0.386$; an adversary inverts it. Selecting
donors by similarity to the \emph{question} makes the three distractors mutually coherent and
leaves the key as the odd one out; a variant selecting donors by similarity to the \emph{key}
inverts the artifact instead of removing it. We could not find a donor rule that lands on
chance.

\paragraph{Model-written distractors may leak to the model that wrote them.} Generating
distractors with Sonnet~4.6, verifying with a second model that each is definitively wrong, and
gating the result yields items on which Sonnet~4.6's blind accuracy is $0.381$
$[0.296, 0.473]$, against $0.274$ $[0.201, 0.363]$ on the human-written originals of the same
113 items. The point estimate moves in the direction one would fear --- the generator's own
family recognising its own writing --- but at $n=113$ the intervals overlap and we cannot call
it established. We report it because the design risk is cheap to avoid: do not generate
distractors with a model from a family you intend to evaluate.

\paragraph{No evidence that negation items drive the effect.} It is natural to suspect items
phrased ``which is \emph{not}\ldots'', where the key is the odd one out by construction.
Sonnet~4.6 scores $0.250$ $[0.138, 0.411]$ blind on the 36 negation items in the pilot and
$0.416$ $[0.343, 0.493]$ on the 161 others. The point estimates point away from the
hypothesis, but 36 items give an interval too wide to separate the two, so this rules the
explanation neither in nor out.

\section{Limitations}

\paragraph{The central claim is a demonstration, not a law.} It rests on one bank and one
gating model: a set filtered against Haiku~4.5 is not clean for GPT-5.6. It does not follow that
every filter fails against every stronger model, only that this one did, decisively, and that
nothing in the procedure prevents it. A practitioner who wants a clean set should gate with
several models from different vendors and accept the lower yield.

\paragraph{What we measured at full scale, and what we did not.} Only the blind condition on
accepted items was re-run on all $8{,}128$; the rejected-set and full-condition columns still
rest on the 600-item sample, with half-widths near $\pm0.05$ and $\pm0.07$. Because every model
was scored on the same 400-item draw, their errors are correlated and the whole column moved
together when we measured the full set --- by $0.010$ on average, which no per-model interval
predicted. The presentation and position checks of \S\ref{sec:ablation}--\S\ref{sec:position}
likewise cover one item set and three models rather than all thirteen.

\paragraph{Coverage is uneven across models.} Llama~3.1~8B returns no parseable letter on
$47.5\%$ of full-condition calls, and a larger token budget does not help; its full-condition
figure rests on 202 of 400 items, which is why it is set apart and excluded from the
correlation. DeepSeek~R1 ($13.1\%$) was budget-bound and recovers at $4{,}096$ tokens. Pixtral~Large's
$5.9\%$ was not a budget effect: in the retained sweep, 67 of its 71 calls without a letter are
throttling errors that the harness recorded as empty replies; of the other four, three are
refusals to answer without the question and one is a Cyrillic letter. Nova~Micro loses $2.4\%$; the other nine models parse above $99.5\%$.

\paragraph{What the design cannot tell us.} We have no human blind baseline, so we cannot say
whether a Ukrainian lawyer shown only the options would score near the pooled $0.383$ we report
for the gating model, which would make the effect a property of the item-writing genre rather
than of models. The capability association rests on eleven models, none below $0.436$
full-condition accuracy, and the claim that item format governs the effect rests on a comparison
of two benchmarks. Extraction fidelity is verified end-to-end on 200 items and by aggregate on
the rest. Our test for recall of the public bank (\S\ref{sec:crossvendor}) finds none behind the excess,
but its control model is weaker than the models it is meant to explain. Finally, GPT-5.6 reaches $0.958$ on the accepted set: for the strongest model tested
the filtered benchmark is close to saturated, and its remaining discriminative value sits in the
models below it.

\section{Conclusion}

Multiple-choice benchmarks should report what a model scores without the question next to what
it scores with it --- and should subtract, from that score, what the model's own
answer-position habit would have earned anyway. The raw blind number is not interpretable
alone. On our gate-accepted set Llama~3.1~8B scores $0.292$ blind --- above every model in Table~\ref{tab:habit} except
the two that actually leak, and three and a half points above DeepSeek~R1 --- entirely because it answers
\texttt{A} to $92\%$ of items and the key sits at \texttt{A} in $29\%$ of them. Corrected for habit, its excess is
$+0.005$, and only two of the original twelve read more than $0.015$ out of the options, joined
by two of five further models (\S\ref{sec:crossvendor}) --- but those read a great deal, GPT-5.6 recovering $0.515$ of a set that was filtered until another model
could not.

The headline of that measurement is what filtering does not buy. Our gate ran until Haiku~4.5
scored $0.204$ on the items it kept; GPT-5.6, which took no part in the selection, answers
$0.515$ of the same items with the question hidden. Filtering against one model does not
neutralise a bank for a better one, and there is no version of the procedure we tested that
does. The surviving items also appear to carry a signature of the model that selected them ---
agreement with its blind choices tracks how far the selection sets another model back --- though
that relationship rests on twelve points and one of them carries it, so we offer it as
an observation to test rather than a result.

Both of these are invisible at the sample sizes ordinarily used. On 400 items nine models read
as ``statistically at chance'', a tidy result that concealed the habit correction and the rest.
Scoring every model on one shared sample also correlates their errors, so the whole column can
move together --- and did, by $0.010$, when we measured the full set.

The same probe on LEXam returns chance, for a reason that is a design choice rather than an
accident: its options point at statements in the stem instead of standing alone. Where options
must be self-contained legal propositions, publish the habit-corrected blind baseline per model
alongside the headline score. Where the format allows reference-style options, the problem does
not arise.

\appendix
\section{Prompts}
\label{app:prompts}

The prompts below are in Ukrainian, matching the language of the bank; the LEXam comparison in
\S\ref{sec:lexam} uses the same two prompts in English, since that split is German and
English. Options are presented as \texttt{A)}--\texttt{D)} in the order given, except in the
gate, where the order is permuted per trial.

\paragraph{Blind condition.} \foreignlanguage{ukrainian}{Нижче наведено чотири варіанти
відповіді на приховане юридичне питання. Питання вам не показано. Оберіть той варіант, який
найімовірніше є правильним. Відповідайте ЛИШЕ однією літерою: A, B, C або D.}

\paragraph{Full condition.} \foreignlanguage{ukrainian}{Питання кваліфікаційного тестування
для кандидатів на посаду судді. Оберіть єдиний правильний варіант. Відповідайте ЛИШЕ однією
літерою: A, B, C або D.}

\paragraph{Blind condition, LEXam.} Below are four answer options to a hidden legal exam
question. The question is not shown to you. Choose the option most likely to be correct. Reply
with ONE letter only: A, B, C or D.

\paragraph{Full condition, LEXam.} Legal exam question. Choose the single correct option.
Reply with ONE letter only: A, B, C or D.

Vendors answer in three shapes --- a bare letter, an explicit
\foreignlanguage{ukrainian}{Відповідь: X} followed by prose, or a paragraph of reasoning
ending in the letter --- so the parser tries an explicit-marker match, then a leading letter,
then the last standalone letter. Unparsed rates are below $0.5\%$ for nine of the thirteen
models. The exceptions, on the 600-item sweep, are DeepSeek~R1 ($13.1\%$), Llama~3.1~8B
($25.5\%$, and $47.5\%$ in the full condition alone), Pixtral~Large ($5.9\%$) and Nova~Micro
($2.4\%$). Unparsed responses are excluded from that model's denominators rather than scored as
wrong. Scoring them as wrong instead, with the habit computed over the same denominator (a response
without a letter earns neither content nor position credit), moves no excess in
Table~\ref{tab:habit} by more than $0.001$. If only the accuracy is charged and the habit is
left on parsed responses, DeepSeek~R1 falls to $+0.010$ $[+0.001, +0.019]$ and Llama~3.1~8B to
$-0.004$ $[-0.008, +0.001]$; neither reading changes which models leak.

The marker match has one known gap. It allows only a colon, dash or emphasis between the marker and
the letter, so a reply ending ``\foreignlanguage{ukrainian}{Відповідь (за загальним правилом): B}''
falls through to the leading-letter rule and is read as whatever letter the reply opens with.
An independent rerun of two of our models on local hardware found one such reply in
600~\citep{indahl2026pinned}. We did not retain raw response text for the main runs, so we cannot
count these cases retrospectively; from revision \texttt{0faa02f} the scripts behind the
full-scale runs accept a parenthetical of up to 80 characters after the marker and take the last
marker in the reply, and the revision accompanying this version brings the 600-item sweep and
the robustness scripts in line. In the $9{,}600$ responses we did retain, from the vendor sweep, the
corrected and original parsers agree on every reply.

\paragraph{A rerun with raw text retained.} In November 2026 we reran the blind condition for
the two Anthropic models on all $8{,}128$ accepted items through the first-party Anthropic API
instead of Bedrock, twice each, with the same prompt, option order, temperature~0 and token
budget, and kept every reply ($32{,}512$ calls). The parser question is settled for these two
models: the corrected and original parsers agree on every one of the $32{,}512$ replies, none is
unparsed, and every reply longer than one character (77 per repeat for Haiku, 104--115 for Sonnet) states the
letter it is read as, in bold, after the marker, or before its option text. Haiku~4.5 answered identically in both repeats on all $8{,}128$ items; Sonnet~4.6
changed letter on $132$ ($1.6\%$). Against the Bedrock run the letters agree on $94.7\%$
(Haiku) and $93.3\%$ (Sonnet) of items, yet no total moves: Haiku scores $0.206$ against
$0.204$, excess $-0.041$ $[-0.050, -0.033]$ against $-0.043$; Sonnet $0.321$ against
$0.320$, excess $+0.083$ $[+0.073, +0.092]$ against $+0.081$; every paired accuracy
difference has an interval spanning zero. The item-level instability is real, at about one
item in fifteen across serving stacks, and it averages out at this sample size
\citep[as also found by][]{indahl2026pinned}.

\section{Models}
\label{app:models}

All models were reached through Amazon Bedrock in August 2026. Snapshot identifiers are given in
full, since several of these aliases will move. The lower half of the conventional group ---
Qwen3~32B, Gemma~3~12B, Ministral~8B, Nova~Micro and Llama~3.1~8B --- was added to extend the
capability range downwards; three of those overlap with the small open-source group evaluated by
\citet{fan2026lexam}, which makes the two model sets partially comparable.

\begin{table}[H]
\centering\small
\caption{Model identifiers, whether the provider accepted a temperature setting, and whether
the model reasons before answering at its default settings. The region column is where the
600-item sweep ran; the full-scale blind run sharded Qwen3, Gemma, Ministral and Pixtral across
several regions to work around per-profile quotas, which is recorded per call in the release. Following
\citet{fan2026lexam}, reasoning and conventional models are grouped rather than equalised:
reasoning cannot be switched off in GPT-5.6, which rejects \texttt{reasoning\_effort} on
Bedrock as it rejects \texttt{temperature}.}
\begin{tabular}{lllcc}
\toprule
\textbf{Model} & \textbf{Bedrock identifier} & \textbf{Region} & \textbf{Greedy} & \textbf{Reasons} \\
\midrule
\multicolumn{5}{@{}l}{\emph{Reasoning models}} \\
GPT-5.6       & \texttt{global.openai.gpt-5.6-sol}                 & eu-central-1 & no  & yes \\
DeepSeek R1   & \texttt{us.deepseek.r1-v1:0}                       & us-east-1    & yes & yes \\
\midrule
\multicolumn{5}{@{}l}{\emph{Conventional models}} \\
Sonnet 4.6    & \texttt{eu.anthropic.claude-sonnet-4-6}            & eu-central-1 & yes & no \\
Haiku 4.5     & \texttt{eu.anthropic.claude-haiku-4-5-20251001-v1:0} & eu-central-1 & yes & no \\
Nova Pro      & \texttt{eu.amazon.nova-pro-v1:0}                   & eu-central-1 & yes & no \\
Nova 2 Lite   & \texttt{eu.amazon.nova-2-lite-v1:0}                & eu-central-1 & yes & no \\
Pixtral Large & \texttt{eu.mistral.pixtral-large-2502-v1:0}        & eu-central-1 & yes & no \\
Llama 3.3 70B & \texttt{us.meta.llama3-3-70b-instruct-v1:0}        & us-east-1    & yes & no \\
Qwen3 32B     & \texttt{qwen.qwen3-32b-v1:0}                       & us-east-1    & yes & no \\
Gemma 3 12B   & \texttt{google.gemma-3-12b-it}                     & us-east-1    & yes & no \\
Ministral 8B  & \texttt{mistral.ministral-3-8b-instruct}           & us-east-1    & yes & no \\
Nova Micro    & \texttt{eu.amazon.nova-micro-v1:0}                 & eu-central-1 & yes & no \\
Llama 3.1 8B  & \texttt{us.meta.llama3-1-8b-instruct-v1:0}         & us-east-1    & yes & no \\
\bottomrule
\end{tabular}
\end{table}

GPT-5.6 rejects the \texttt{temperature} parameter outright, so its requests omit it; every
other model was run at temperature 0. The 600-item sweep used a 2{,}048-token output budget and
the full-scale blind run 4{,}096; the budget matters for two models for opposite reasons. Nova Pro is merely verbose: it writes a paragraph
of justification around the letter, some 260 output tokens of which nearly all is visible text.
DeepSeek~R1 spends its tokens on a reasoning trace the response body does not contain,
averaging $1{,}363$ output tokens per blind item over the full-scale run for a handful of
visible characters. It is the model the budget bound hardest --- at 2{,}048 tokens it failed to
reach a letter on $13.1\%$ of calls, at 4{,}096 on $1.9\%$, the estimates moving only within
their intervals --- and the only one: Pixtral~Large also fell, from $5.9\%$ to $1.2\%$, but its losses at
2{,}048 tokens were mostly throttling errors recorded as unparsed, not truncation. Because the set mixes models that reason at inference time with models
that do not, \S\ref{sec:reasoning} measures what that mixture is worth rather than leaving it as
a caveat.

{\sloppy The November 2026 runs of \S\ref{sec:crossvendor} used first-party endpoints: Google Vertex AI
for \texttt{gemini-3.1-pro-preview} (\texttt{thinkingLevel} \textsc{low}; default thinking for the
600-item check), and Azure AI Foundry deployments of \texttt{gpt-5.4} (snapshot 2026-03-05),
\texttt{DeepSeek-V4-Pro} (version 2026-04-23), \texttt{Mistral-Large-3}, \texttt{gpt-oss-120b}
and \texttt{gpt-4o} (2024-11-20). All six accepted temperature~0. Unlike the August harness,
these runs record transport errors separately from replies and retry them, and record which
parser rule produced each letter. Across the five models on $8{,}128$ items, gpt-oss-120b alone
lost letters: three replies cut off at the token limit and four calls blocked by Azure's
content filter, seven of $8{,}128$. The last-standalone-letter rule fired for Mistral~Large~3
on 152 replies of the form
``\foreignlanguage{ukrainian}{Найімовірніше правильний варіант:} \textbf{B}'', all of which
we checked against the emphasised letter. Gemini~3.1~Pro at default thinking exhausted the
budget on 20 of 600 items, which we exclude rather than parse from a truncated trace; of its 580
complete replies one needed that rule, and it reads correctly.\par}

\section*{Data and Code}

The corpus (11{,}990 items), the gated subset (8{,}128), the 600-item cross-vendor sample,
all 9{,}600 blind and full predictions from the first sweep and 7{,}200 from the small-model
extension, the negative-result runs, the $21{,}600$ calls of prompt, labelling and position
ablation, the reasoning controls, the $105{,}664$-call full-scale blind run, the November 2026
cross-vendor, control and memorisation runs with every reply, the re-extracted text, and the
extraction, verification and gating code are released at
\url{https://huggingface.co/datasets/overthelex/ua-judge-exam}.

\bibliographystyle{plainnat}
\bibliography{references}

\end{document}